\documentclass[lettersize,journal]{IEEEtran}
\usepackage{amsmath,amsfonts}
\usepackage{algorithmic}
\usepackage{algorithm}
\usepackage{array}
\usepackage[caption=false,font=normalsize,labelfont=sf,textfont=sf]{subfig}
\usepackage{textcomp}
\usepackage{stfloats}
\usepackage{url}
\usepackage{verbatim}
\usepackage{graphicx}
\usepackage{cite}
\usepackage{upgreek}
\usepackage[colorlinks=true, allcolors=blue]{hyperref}
\usepackage{placeins}
\usepackage{bm}

\newcommand{\AlgorithmicInput}[1]{\item[\textbf{Input:}] {#1}\\}
\newcommand{\AlgorithmicOutput}[1]{\item[\textbf{Output:}] {#1}\\}

\begin{document}

\title{Task Delay and Energy Consumption Minimization for Low-altitude MEC via Evolutionary Multi-objective Deep Reinforcement Learning}

\author{Geng Sun,~\IEEEmembership{Senior Member,~IEEE,} 
        Weilong Ma, 
        Jiahui Li, 
        Zemin Sun, 
        Jiacheng Wang,\\
        Dusit Niyato,~\IEEEmembership{Fellow,~IEEE},
        Shiwen Mao,~\IEEEmembership{Fellow,~IEEE}

    \thanks{This work is supported in part by the National Natural Science Foundation of China (62272194, 62471200), in part by the Science and Technology Development Plan Project of Jilin Province (20230201087GX), in part by the Postdoctoral Fellowship Program of China Postdoctoral Science Foundation (GZC20240592), in part by China Postdoctoral Science Foundation General Fund (2024M761123), and in part by the Scientific Research Project of Jilin Provincial Department of Education (JJKH20250117KJ). \textit{(Corresponding author: Jiahui Li.)}}
    \thanks{Geng Sun is with the College of Computer Science and Technology, Jilin University, Changchun 130012, China, and with Key Laboratory of Symbolic Computation and Knowledge Engineering of Ministry of Education, Jilin University, Changchun 130012, China; he is also affiliated with the College of Computing and Data Science, Nanyang Technological University, Singapore 639798 (e-mail: sungeng@jlu.edu.cn).}
    \thanks{Weilong Ma is with the College of Software, Jilin University, Changchun 130012, China (e-mail: wlma22@jlu.edu.cn).}
    \thanks{Jiahui Li and Zemin Sun are with the College of Computer Science and Technology, Jilin University, Changchun 130012, China (e-mails: lijiahui@jlu.edu.cn; sunzemin@jlu.edu.cn).}
    \thanks{Jiacheng Wang and Dusit Niyato are with the College of Computing and Data Science, Nanyang Technological University, Singapore (e-mails: jiacheng.wang@ntu.edu.sg; dniyato@ntu.edu.sg). }
    \thanks{Shiwen Mao is with the Department of Electrical and Computer Engineering, Auburn University, Auburn, AL 36849-5201 USA (e-mail: smao@ieee.org).}
}



\maketitle
\begin{abstract} 
The low-altitude economy (LAE), driven by unmanned aerial vehicles (UAVs) and other aircraft, has revolutionized fields such as transportation, agriculture, and environmental monitoring. In the upcoming six-generation (6G) era, UAV-assisted mobile edge computing (MEC) is particularly crucial in challenging environments such as mountainous or disaster-stricken areas. The computation task offloading problem is one of the key issues in UAV-assisted MEC, primarily addressing the trade-off between minimizing the task delay and the energy consumption of the UAV. In this paper, we consider a UAV-assisted MEC system where the UAV carries the edge servers to facilitate task offloading for ground devices (GDs), and formulate a calculation delay and energy consumption multi-objective optimization problem (CDECMOP) to simultaneously improve the performance and reduce the cost of the system. Then, by modeling the formulated problem as a multi-objective Markov decision process (MOMDP), we propose a multi-objective deep reinforcement learning (DRL) algorithm within an evolutionary framework to dynamically adjust the weights and obtain non-dominated policies. Moreover, to ensure stable convergence and improve performance, we incorporate a target distribution learning (TDL) algorithm. Simulation results demonstrate that the proposed algorithm can better balance multiple optimization objectives and obtain superior non-dominated solutions compared to other methods.
\end{abstract}

\section{Introduction}
\label{sec:introduction}

\par \IEEEPARstart{W}{ith} the rapid development of the six-generation (6G) and internet-of-things (IoTs), a wide range of compelling applications have emerged. However, the computational resources of these ground devices (GDs) are often constrained by production costs and device size, thereby affecting the computational performance and user experience. Under such circumstances, mobile edge computing (MEC) arises as an effective approach by offloading the computational resources to edge servers, and it has been shown to significantly enhance the capabilities of GDs in performing computationally intensive tasks \cite{Feng2022}. According to the MEC paradigm, GDs offload tasks to the nearby base stations (BSs), so that reducing the task processing delay and lowering the requirements for the configuration and energy consumption of GDs. However, in many cases, it is infeasible for the fixed BSs to cover all GDs, thereby resulting in incomplete task execution.

\par The emergence of the low-altitude economy (LAE) has further accelerated the use of unmanned aerial vehicles (UAVs) in wireless communication and computation systems. As an economic model driven by low-altitude flight activities, LAE integrates UAVs into applications such as urban logistics, precision agriculture, and environmental surveillance, creating significant economic and societal value. As such, the paradigm of UAV-assisted communication is anticipated to be essential for MEC systems \cite{Wu2021}. This is due to the fact that UAVs can offer more flexibility than the fixed BSs, and it is able to better support various mobile services \cite{Ning2023}. As such, the concept of flying MEC was proposed, which further considers the inclusion of computing resources where the UAVs can carry MEC servers \cite{Du2018}. Specifically, flying MEC combines the advantages of UAVs with high flexibility and speed, and may provide higher-quality computing services compared to traditional MEC infrastructure\cite{Gao2024}.

\par Different from previous studies, this paper aims to propose a dynamic optimization strategy and find reasonable balances between the task delay and the energy consumption of the UAV in UAV-assisted MEC systems of LAE scenarios. However, achieving such goals in LAE is a challenging task. First, the time cost of task offloading is related to the speed control of the UAV. Thus, task offloading and trajectory planning are mutually coupled and jointly affect the two aforementioned objectives, which are challenging to determine. Second, these two objectives are conflicting and their relative importance may vary in different scenarios. As such, some existing optimization methods \cite{Ye2020} that consider only a single optimization objective and add others as constraints, or that combine multiple goals into one optimization objective, are less generalizable. Therefore, we aim to optimize these mutually coupled variables to find several universally applicable and balanced policies. The main contributions of this paper are summarized as follows.

\begin{itemize}
     \item \textit{Dual-objective Problem Formulation in UAV-assisted MEC System:} We investigate a UAV-assisted MEC system where the UAV carries edge servers to facilitate the computation services and task scheduling for GDs. This system involves complex optimization challenges for balancing the task delay and the energy consumption of the UAV, including task offloading decisions and UAV trajectory planning. As such, we formulate a calculation delay and energy consumption multi-objective optimization problem (CDECMOP) and seek to optimize the inherent conflict between the costs of task delay and energy consumption.
    
    \item \textit{Evolutionary Multi-Objective DRL Algorithm:} In the LAE scenarios, our CDECMOP faces a complex long-term scheduling problem, which is challenging to address directly by using the traditional deep reinforcement learning (DRL) methods. To overcome these challenges, we first employe a scheduling method based on simulated annealing (SA) to simplify the potential action space and reformulated it as a multi-objective Markov decision process (MOMDP). To effectively handle the formulated MOMDP, we employ an evolutionary multi-objective DRL (EMODRL) framework. Specifically, we utilize a multi-strategy proximal policy optimization (PPO) to enable the EMODRL and introduce a multi-objective target distribution learning (TDL) algorithm to optimize the offspring population update mechanism in EMODRL, which stabilizes the evolution process at each step, ultimately leading to a superior Pareto-optimal strategy.
    
    \item \textit{Simulation and Performance Evaluation:} Simulation results demonstrate that the proposed algorithm achieves a high-quality set of Pareto solutions. Additionally, extensive comparisons show that the strategy obtained by the proposed algorithm significantly outperforms various baseline methods. Moreover, it is also found that the proposed algorithm surpasses other EMODRL methods in terms of both the convergence and effectiveness of non-dominated solutions. 
\end{itemize}

\par The rest of this paper is arranged as follows. Section~\ref{sec:related_work} gives some related works. Section~\ref{sec:models_and_preliminaries} presents the models and preliminaries. Section~\ref{sec:problem_formulation} formulates the CDECMOP of UAV-assisted MEC. Section~\ref{sec:algorithm} presents the MOMDP, and proposes the corresponding algorithm. Simulation results are presented in Section~\ref{sec:simulation_results}. Finally, the paper is concluded in Section~\ref{sec:conclusion}.

\section{Related Work}
\label{sec:related_work}
\par In this section, some related works that explore key challenges in UAV-assisted MEC networks are briefly reviewed.
\par Previous studies on UAV-assisted MEC have shown that the UAVs can provide extensive coverage and reliable line-of-sight (LoS) links for GDs, along with additional computational capabilities. For instance, Zeng \textit{et al.}~\cite{Zeng2016} highlighted the ability of UAVs to significantly enhance mobile communication performance. Yu \textit{et al.}~\cite{Yu2020} discussed how mobile users offload computational tasks from their devices to MEC servers for processing, primarily focusing on task offloading and computational resource allocation. Tran \textit{et al.}~\cite{Tran2019} investigated UAV networks equipped with MEC servers, where the optimization of LoS link quality played a crucial role in enabling efficient task offloading and resource allocation, thereby improving task completion time and energy efficiency. Additionally, multiple studies have highlighted the challenges and issues in MEC and UAV-assisted MEC. Ye \textit{et al.}~\cite{Ye2020} explored the integration of ground-satellite networks for flight MEC, analyzing its advantages and open issues, particularly the complex models of interstellar and space-ground links in terms of energy costs and time. Nguyen \textit{et al.}~\cite{Nguyen2020} proposed an optimization algorithm to schedule the flight speed of UAVs to minimize energy consumption while assisting ground IoT devices in completing computational tasks, and designed a novel graphical method to visualize and analyze this problem. Liu \textit{et al.}~\cite{Liu2020} proposed a collaborative offloading scheme and resource management strategy, using UAV-assisted MEC technology to provide services for power IoT systems, aiming to maximize long-term network utility through DRL algorithms. As can be seen, the current studies focused on single-objective optimization for UAV-assisted MEC task offloading. To facilitate analysis, we introduce them separately according to different optimization objectives.

\par \textit{Task Delay:} The delay of tasks within the system is an important factor that was considered in many studies. For example, Li \textit{et al.}~\cite{Li2020} proposed several trajectory optimization algorithms while considering the wireless resource management and task scheduling to reduce the overall task delay. Moreover, Ren \textit{et al.}~\cite{Ren2022} developed a hierarchical DRL algorithm in a dynamic MEC environment to minimize the average task delay, thereby reducing the computational complexity and improving the convergence efficiency while achieving near-optimal performance. Further, Mach et al.~\cite{Mach2024} proposed a hybrid half and full duplex relaying strategy to effectively reduce task delay in UAV-assisted MEC systems. In addition, some studies have explored diverse approaches for delay minimization in UAV-assisted MEC systems through DRL and blockchain techniques. Wang \textit{et al.}~\cite{Wang2024_multi} proposed an optimization problem for UAV-assisted MEC networks combined with blockchain technology, which integrates UAV positioning, data offloading, and resource allocation to minimize the total time consumption for data processing. Similarly, Zheng \textit{et al.}~\cite{Zheng2024} introduced a DRL-based optimization algorithm to minimize the overall delay in UAV-assisted MEC systems, thereby optimizing task execution efficiency. Furthermore, Wang \textit{et al.}~\cite{Wang2024_Latency} designed a DRL-based framework for joint task offloading and resource allocation, effectively reducing the total task processing delay in multi-UAV-assisted air-ground collaborative MEC systems.

\par \textit{Energy Efficiency:} Some works aimed to minimize the energy consumption of the UAV or the entire system. For example, Sun \textit{et al.}~\cite{Sun2021} proposed that the UAV equipped with computing servers to assist smart devices in processing computational tasks, and they developed a convex optimization approach to improve UAV decision-making, aiming to reduce the UAV energy consumption. Similarly, Ye \textit{et al.}~\cite{Ye2020} investigated the energy-efficient flight speed scheduling problem in an MEC system and proposed a heuristic approach to control the flight of UAV, which achieved near-optimal solutions to the problem. Moreover, Lin \textit{et al.}~\cite{lin2023pddqnlp} proposed a DRL approach to maximize the energy efficiency of UAV. In addition, some studies have also addressed energy-related optimization problems beyond mere energy consumption through DRL. Shi \textit{et al.}~\cite{Shi2024} designed a method based on DRL to optimize trajectories and frequencies in energy-constrained multi-UAV-assisted MEC systems, thereby achieving efficient energy utilization. Similarly, Kim \textit{et al.}~\cite{Kim2024} developed a collaborative multi-agent DRL approach for UAV-assisted MEC networks, which significantly enhances the overall system energy efficiency. Furthermore, Song \textit{et al.}~\cite{Song2024} explored the energy-efficient path planning problem for UAV-assisted MEC with wireless charging, aiming to reduce energy consumption while ensuring effective task execution.

\par In real-world UAV-assisted MEC systems, due to variations in specific scenarios and tasks, considering one single optimization objective may be insufficient. Instead, multiple conflicting objectives often exist. As such, a balance between task delay and the energy consumption of the UAV should be considered during the UAV flight control process by combining the two important optimization objectives. For instance, in~\cite{Wang2020}, the authors established a model to assist the agent in devising the best offloading strategy, ultimately reducing both task completion time and the energy use of the UAV. In~\cite{Mousa2022}, the authors introduced an improved evolutionary metaheuristic algorithm that optimizes the computation offloading strategy, achieving the goal of minimizing task delay and the energy consumption of the UAV. In~\cite{Huang2024},  the authors considered the use of UAV-assisted MEC to provide computational services for resource-limited devices, while employing DRL methods to jointly optimize task processing delay, UAV energy consumption, and the total amount of collected task data. Moreover, in~\cite{Chen2022}, the authors developed an algorithm based on DRL and transfer learning to reduce the delay and energy consumption. In addition, some research tackled UAV control problems with multiple objectives by integrating weights with DRL, such as~\cite{Ke2024}.

\section{Models and Preliminaries}
\label{sec:models_and_preliminaries}

\par In this section, we present the models and some preliminaries. Moreover, we provide a summary of the notations utilized in this paper in Table~\ref{tab:notation}.

\begin{table}[!t]
\caption{Main notations \label{tab:notation}}
\centering
\begin{tabular}{c c}
\hline
Notation                        & Definition                                               \\ \hline
$a_i(t)$                        & The decision of the UAV for GD $i$ at time slot $t$      \\
$B^{\text{UL}}$                 & The bandwidth of the UAV                                 \\
$c_{\text{light}}$              & The speed of light                                       \\
$d_i(t)$                        & UAV distance to GD $i$ at time $t$                       \\
$d_t$                           & The UAV travel distance in time $t$          \\
$d^{h}_{\text{max}}$            & The UAV maximum coverage range               \\
$d_{\text{max}}$                & The max distance of the UAV per time slot                \\
$H$                             & Fixed flying altitude of the UAV                         \\
$\mathcal{I}_t$                 & The set of GDs chosen by the UAV in time slot $t$        \\
$N$                             & The number of GDs                                        \\
$\mathcal{N}$                   & The set of GDs                                           \\
$O_i$                           & The size of the data for task $i$              \\
$\mathcal{P}_{i}^{\text{tran}}$ & The transmit power of GD $i$                             \\
$R_i^{\text{UL}}$               & Uplink rate from GD $i$ to UAV                           \\
$T$                             & Number of time slots                                     \\
$\mathcal{T}$                   & Set of time slots                                        \\
$\delta_{i}(t)$                 & LoS probability for UAV and GD $i$ during time slot $t$      \\
$\eta_{\text{LoS}}$             & The losses for LoS                                       \\
$\eta_{\text{NLoS}}$            & The losses for non-LoS                                   \\
$\theta_t$                      & The horizontal direction of the UAV in time slot $t$     \\
$\theta^{\text{max}}$           & The maximal azimuth angle of UAV                         \\
$\upkappa$                      & The effective switched capacitance                       \\
$\lambda_i$                     & Task creation time                                       \\
$\lambda_i(t)$                  & Path loss for GD $i$ and UAV at time slot $t$            \\
$\mu_i$                         & CPU cycles required for process one bit data            \\
$\sigma^2$                      & The power spectral density of noise                      \\
$\tau$                          & The duration of per time slot                           \\
$\Omega$                        & All the invalid states                                   \\ \hline
\end{tabular}

\end{table}
\subsection{System Overview}
\begin{figure}[t]
    \centering
    \includegraphics[scale=0.4]{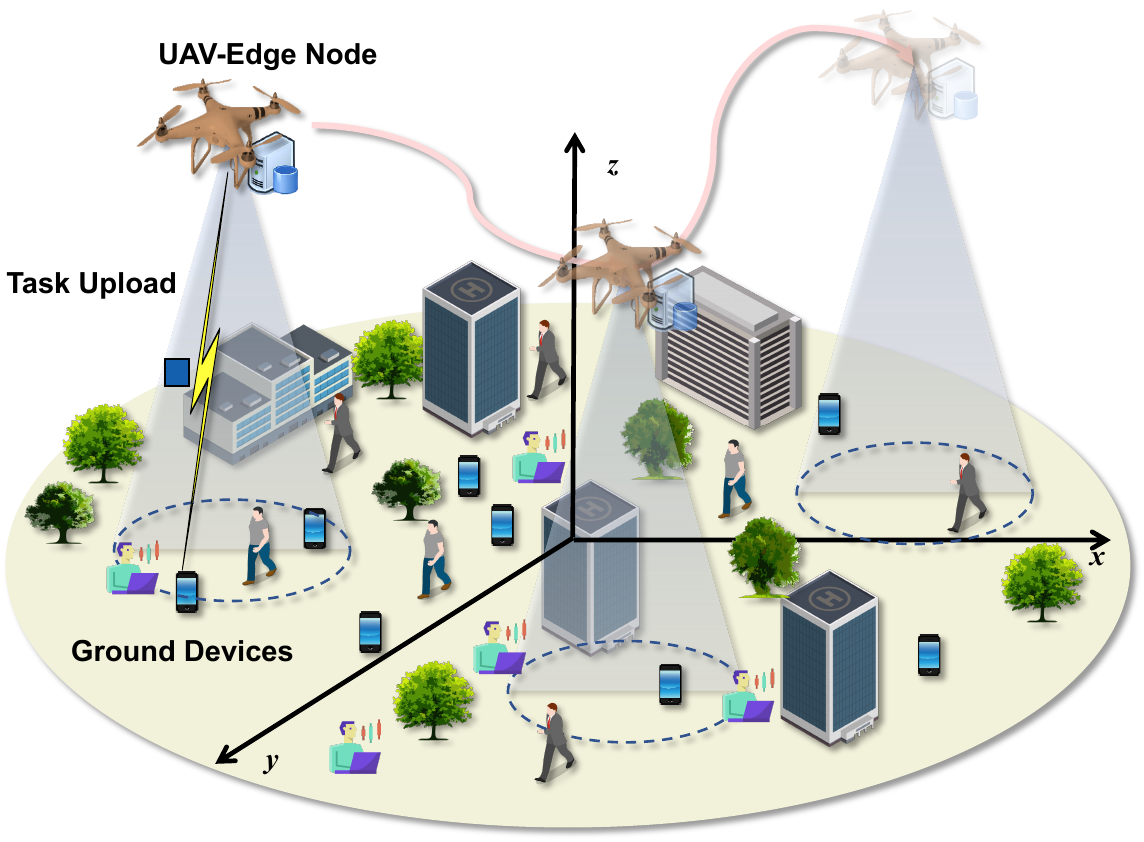}
    \caption{A schematic diagram of the system for task computation assisted by a UAV equipped with an edge server.}
    \label{fig:system_model}
\end{figure}
\par As depicted in Fig.~\ref{fig:system_model}, we consider a UAV-assisted MEC system that includes a UAV-carried the edge servers and multiple GDs. Specifically, the GDs, denoted as $\mathcal{N} = \{1, \ldots, N\}$, are randomly distributed within an area and follow a uniform distribution. Moreover, each GD dynamically generates computational tasks, and these tasks are predominantly computation-intensive and time-sensitive, indicating that the GDs which may not have sufficient computing resources, cannot complete these computational tasks locally. In such cases, a UAV-carried edge servers is dispatched to receive and execute computational tasks from the GDs dynamically. To enhance the stability of this process, the UAV caches the tasks of the GDs in a queue and executes them in an orderly manner.

\par Without loss of generality, we divide the total maximum task processing time $T$ into a discrete-time system. Each segment is referred to as a time slot with a duration of $\tau$, and the entire time set can be expressed as $\mathcal{T} = \{1, \ldots, T\}$. Moreover, the proposed system is established on a three-dimensional Cartesian coordinate system. As such, the positions of the $i$th GD and the UAV are denoted by $w_i = (x_i^{\text{GD}}, y_i^{\text{GD}}, 0) \in \mathbb{R}^3$ and $q_{t}^{U} = (x_t^{U}, y_t^{U}, H) \in \mathbb{R}^3$, $1 \leq t \leq T$, respectively. Specifically, $(x_i^{\text{GD}}, y_i^{\text{GD}})$ represents the horizontal coordinates of the $i$th GD, $(x_t^{U}, y_t^{U})$ represents the horizontal coordinates of the UAV, $H$ represents the altitude of the UAV, and $T$ represents the duration of the mission.

\par Accordingly, the horizontal distance $d_{i}^{h}(t)$ between the UAV and GD $i$ at time slot $t$ can be calculated as follows:
\begin{equation}
d_{i}^{h}(t) = \sqrt{(x_t^{U} - x_i^{\text{GD}})^2 + (y_t^{U} - y_i^{\text{GD}})^2}\text{.}
\end{equation}
\par Furthermore,  we define $\theta^{\text{max}}$ as the maximum azimuth angle of the UAV. Accordingly, $d^{h}_{max}=H \cdot \tan(\theta^{\text{max}})$ represents the UAV maximum coverage range. Following this, the UAV can communicate with GD $i$ and perform task upload only if GD $i$ falls within the coverage range of the UAV, defined by $d_{i}^{h}(t) \leq d^{h}_\text{max}$.

\subsection{Task Model}

\par We consider that the collaboration among all GDs is allowed for sharing the basic information such as task generation timelines. Accordingly, if GD $i$ generates a task, a broadcast message is disseminated through the networks of the GDs. Consequently, the UAV can easily obtain the task-generation information from the GDs within its coverage, thereby utilizing it for the subsequent decision-making processes.
\par Following this, the computational tasks generated by GDs are considered to be computation-intensive and time-sensitive. Specifically, each GD $i$ periodically handles computation-intensive tasks $\Psi_i = \langle O_i, \mu_i, \lambda_i \rangle$, where $O_i$ represents the size of the task data, $\mu_i$ denotes the amount of computing resources required for processing a unit of data, and $\lambda_i$ indicates the task arrival time. Note that all the generated but unfinished tasks remain in a waiting status from their creation until they receive a message indicating that the UAV is ready to accept tasks. Upon receiving this message, the GDs upload the tasks to the task queue in the UAV. The connection between the $i$th GD and the UAV at time slot $t$ is given as follows:
\begin{equation}
    a_i(t) = \{0, 1\}, \forall i \in \mathcal{N}, t \in \mathcal{T} \text{,}
\end{equation}
\noindent where $a_i(t) = 1$ means that the UAV chooses to receive the task $i$, and vice versa.
\par Note that the UAV successfully collects tasks from the GD only when the GD is within its coverage. Thus, the successful task upload must satisfy the conditions as follows:
\begin{equation}
    a_i(t) \cdot d_{i}^{h}(t) \leq d^{h}_\text{max}, \forall i \in \mathcal{N}, t \in \mathcal{T} \text{.}
\end{equation}
\subsection{Communication Model}

\par During the communication process, the task transmission from the GDs to UAVs incurs delays in handling tasks, and we model this process and analyze the corresponding delays as follows.
\par We begin by presenting the channel fading model. Specifically, the transmission distance from the UAV to GD $i$ is represented as $d_{i}(t) = \sqrt{(x_t^{U} - x_i^{\text{GD}})^2 + (y_t^{U} - y_i^{\text{GD}})^2 + H^2}$, and then the LoS probability between GD $i$ and the UAV in time slot $t$ can be expressed as follows ~\cite{Mei2020}:
\begin{equation}
\begin{aligned}
\delta_{i}(t) = \frac{1}{1 + a \cdot \exp { \left(-b\left(\arctan\left(\frac{H}{d_{i}(t)}\right)-a\right) \right)}} 
\end{aligned}
\text{,}
\end{equation}
\noindent where $a$ and $b$ are the constants determined by the environment. Thus, the path loss between GD $i$ and the UAV at time slot $t$ can be expressed as follows:
\begin{equation}
\begin{aligned}
\lambda_{i}(t) = &20\log(\sqrt{H^2+d^2_{i}(t)})
+\delta_{n}(t)(\eta_{\text{LoS}}-\eta_{\text{NLoS}})\\ 
&+20\log\left(\frac{4\pi f}{c_{\text{light}}}\right)+\eta_{\text{NLoS}} \text{,}
\end{aligned}
\end{equation}
\noindent where $f$ represents the carrier frequency, and $\eta_{\text{LoS}}$ and $\eta_{\text{NLoS}}$ correspond to the losses for LoS and non-LoS, respectively. Given that all the links share a common frequency band for multiplexing, the signal-to-interference-and-noise ratio (SINR) at the UAV during the uplink communication from GD $i$ at time slot $t$ can be described as follows:
\begin{equation} 
\begin{aligned} 
\gamma_{i}(t) = 
    \cfrac{v_i(t)\mathcal{P}^{\text{tran}}_{i}10^{-\lambda_{i}(t)/10}}{\sum_{j=1}^N v_j(t)\mathcal{P}^{\text{tran}}_{j}10^{-\lambda_{j}(t)/10}+\sigma^2} 
\end{aligned}
\text{,}
\end{equation}
\noindent where $\mathcal{P}_{i}^{\text{tran}}$ represents the transmit power of GD $i$, and $\sigma^2$ denotes the power spectral density of noise. Moreover, $v_{i}(t)\in \{0, 1\}$ indicates whether GD $i$ is communicating with the UAV.

\par Therefore, the achievable uplink data transmission rate from GD $i$ to the UAV within time slot $t$ can be calculated as follows:
\begin{equation}
\begin{aligned}
    R_i^{\text{UL}} = B^{\text{UL}}\log_2 \left(
                    1 + \gamma_i(t)
                            \right)
\end{aligned}
\text{,}
\end{equation}
\noindent where $B^{\text{UL}}$ represents the bandwidth of the UAV.

\par Note that all tasks wait for the selection notification of the UAV. Specifically, when the tasks are selected and within the coverage of the UAV, they are offloaded to the UAV through the ground–to-air (G2A) link without any local calculation. Moreover, the delay for task $i$ of GD $i$ at time slot $t$ waiting to enter the UAV is recorded as $D_i^{\text{wait}}(t)$, and the transmission delay of task $i$ to the UAV is given as follows:
\begin{equation}
D_i^{\text{G2A}} =
\cfrac{O_i}{R_i^{\text{UL}}} \text{.}
\end{equation}

\subsection{Computational Model}

\par After the tasks are transmitted to the UAV, they are scheduled within the task queue of the UAV. We present the delay and energy consumption incurred by the task computation within the UAV in this subsection.
\par When the computing task from GD $i$ enters the task queue of the UAV, the UAV schedules the tasks to minimize the total task delay. Specifically, the time from entering the queue to being scheduled is denoted as $D_i^{s}$, and once the task is scheduled, the UAV begins the computation. The delay in processing the task offloaded from GD $i$ is given as follows:
\begin{equation}
D_i^{U} =
\cfrac{O_i \cdot \mu_i}{f_{i}^{U}} \text{,}
\end{equation}

\noindent where $f_i^{U}$ (measured in CPU cycles/s) represents the computational resources assigned by the UAV to GD $i$. 
\par Correspondingly, the energy consumption arising from task execution is separated into two specific parts. The first part is the energy consumption generated by task computation, which depends on the power consumption of the CPU and the computation time. Therefore, the energy consumption generated by the UAV processing task $i$ can be expressed as follows:
\begin{equation}
E_i^{\text{CP}} 
=\upkappa (f_i^{U})^3D_i^{U}
=\upkappa O_i\mu_i \left( f_{i}^{U} \right) ^2 \text{,}
\end{equation}
\noindent where $\upkappa$ is a constant that depends on the CPU architecture.
\par The second part is the transmission energy consumption generated by the UAV receiving the task $i$, which is given as follows:
\begin{equation}
E_i^{\text{RX}} 
=P_{\text{RX}}^{U}D_i^{\text{G2A}}
=\cfrac{O_i P_{\text{RX}}^{U}}{R_i^{\text{UL}}}
\text{,}
\end{equation}
\noindent where $P_{\text{RX}}^{U}$ represents the UAV receiving power.

\subsection{UAV Movement Model}

\par In this subsection, we detail the movement model of the UAV. For simplicity, we consider that the UAV flies at a constant altitude denoted by $H$, and the horizontal direction and distance traveled by the UAV in time slot $t$ are denoted by $\theta_t$ and $d_t$, respectively. Then, due to the power budget constraint, the UAV has a maximum flight speed of $v_{\text{max}}$. Therefore, the maximum flight distance of the UAV within a single time slot is $d_{\text{max}}=v_{\text{max}} \times \tau$. 

\par We employ the Cartesian coordinate system for describing the UAV position at a given time. Specifically, let $q_t^h = (x_t^U, y_t^U)$ represent the horizontal position of the UAV in time slot $t$. Subsequently, the UAV moves a distance $d_t$ at an angle $\theta_t$, and thus its position at time slot $t+1$ is updated as follows ~\cite{ji2023}:
\begin{equation}
    \begin{cases}
        x_{t+1}^{U}=x_{t}^{U}+d_t \cdot \cos(\theta_t) \\
        y_{t+1}^{U}=y_{t}^{U}+d_t \cdot \sin(\theta_t) 
    \end{cases} 
    \text{.}
\end{equation}

\par Additionally, when the UAV operates at the speed of $v_t = d_t / \tau$ during time slot $t$, the corresponding propulsion power consumption $P(v_t)$ of UAV is given as follows:
\begin{equation} 
    \begin{aligned}
    P(v_t) = &P_1\left(
                    1+\cfrac{3v_t^2}{v_{\text{tip}}^{2}}
                \right) + 
            P_2 \left(
                \sqrt{1 + \cfrac{v_t^4}{4v_0^4}} - \cfrac{v_t^2}{2v_0^2}
                \right)^{\frac{1}{2}}\\
            &+ \cfrac{1}{2}d_0\rho sAv^3  ,
    \end{aligned}
\end{equation}
\noindent where $P_1$ and $P_2$ are constants associated with blade profile power and induced power during hover, respectively, and $v_0$ and $v_{\text{tip}}$ are constants for the average rotor induced speed in hover and the rotor blade tip speed. Moreover, $A$ and $s$ represent the rotor disc area and the rotor solidity, respectively, $\rho$ and $d_0$ denote the air density and the fuselage drag ratio.

\par Then, the energy consumption $E_{\text{fly}}$ caused by UAV flight is given as follows:
\begin{equation}
    E_{\text{fly}} = \int_{0}^{T}P(v_t)dt \text{.}
\end{equation}

\section{Problem Formulation}
\label{sec:problem_formulation}

\par The primary objectives of the considered system are to minimize the total task delay and the energy consumption of the UAV. Due to the need for tasks to be completed quickly, the UAV may need to fly quickly to the vicinity of the GD generating the task. However, frequent movement of the UAV leads to the increased energy consumption. Similarly, it is crucial for the UAV to execute GD tasks in a timely manner. However, increasing the task handling speed will significantly escalate the computational load on the UAV, thereby leading to an increase in UAV energy consumption. Thus, it is crucial to jointly control the real-time flight direction and speed of the UAV, optimize task upload decisions at each time slot, manage the scheduling sequence of tasks in the queue, and find balanced policies that can balance these conflicting objectives.

\par Consequently, we first detail the optimization objectives, and then formulate the corresponding CDECMOP.

\par \textit{Total Task Delay:} In the considered scenario, we consider that due to the limited computational capacity of the GDs, they do not perform local computation and instead offload their computational tasks to the UAV for processing. Note that the output data size of the computation results is often much smaller than the input data size in many computation-intensive applications. Therefore, the time required to send the computation results back to GDs can be ignored~\cite{Hu2019}. In this case, the total delay during the task processing comprises three main parts, detailed as follows: (a) Waiting time from task generation to offloading to the UAV. (b) The task transmission delay between GDs and the UAV. (c) The delay experienced by tasks after entering the UAV, while awaiting scheduling and execution. 

\par As such, considering that the UAV selects the service GD $i$ in the time slot $t$, and the set of all GDs selected in this time slot is $\mathcal{I}_t$, then the calculation delay during the time slot $t$ is expressed as follows:
\begin{equation}
    D_{t} = \sum_{i \in \mathcal{I}_t} 
            \left(
                D_i^{U}+
                D_i^{\text{G2A}}
            \right)
    \text{.}
\end{equation}

\par Accordingly, the first optimization objective is to minimize the total task delay, which can be given as follows:
\begin{equation}
f_{1} = \sum_{t = 1}^{T} (D_t
+ \sum_{i = 1} ^ {N} D_i^{\text{wait}}(t)) + \sum_{i = 1} ^ {N}D_i^{s}
\text{,}
\label{eq:f1}
\end{equation}
\noindent where $D_i^{\text{wait}}$ represents the waiting time for a task from its generation until it is uploaded to the UAV, and $D_i^s$ represents the scheduling time for the task within the UAV.

\par \textit{Total Energy Consumption of the UAV:} The total energy consumption during task processing is divided into two main parts, which are the energy consumption of the UAV caused by the CPU processing the task and the energy consumption generated by the UAV during flight. Consequently, the energy consumption of the UAV for computing and receiving tasks from the selected service of GD $i$ during the $t$-th time slot is expressed as follows:
\begin{equation}
    E_{t} = \sum_{i \in \mathcal{I}_t} 
            \left(
                E_i^{\text{CP}}+
                E_i^{\text{RX}}
            \right)
    \text{.}
\end{equation}

\par Thus, the second optimization objective is to minimize such energy consumption of the UAV, which is given as follows:
\begin{equation}
    f_{2} = \sum_{t = 1}^{T} 
            E_t
            + E_{\text{fly}}
    \text{.}
\end{equation}
\par Accordingly, we formulate a CDECMOP to simultaneously minimize the total task delay (\textit{i.e.}, $f_{1}$) and the total energy consumption of the UAV (\textit{i.e.}, $f_{2}$) by jointly optimizing the flying trajectory of the UAV (\textit{i.e.}, $\theta_t$ and $d_t$) and the selection decision for each task (\textit{i.e.}, $a_i(t)$), which is as follows:
\begin{subequations}
\label{eq:formu}
\begin{align} 
    \mathop{\text{min}}\limits_{\{\theta_t, d_t, a_i(t)\}} F &=
        \left\{
            f_{1}, f_{2}
        \right\} \text{,} \label{eq:moo-fomulate:a} \\
\text{s.t. }
    &\theta_{t} \in [0, 2 \pi], \forall t \in \mathcal{T} \text{,}  \label{eq:moo-fomulate:b}\\
    &0 \leq d_{t} \leq d_{\max }, \forall t \in \mathcal{T} \text{,}  \label{eq:moo-fomulate:c} \\
    &a_{i}(t) \in \{0, 1\}, \forall i \in \mathcal{N} \text{, }\forall t \in \mathcal{T} \text{,}  \label{eq:moo-fomulate:d} \\
    &0 \leq x_{t}^{\mathrm{U}} \leq x_{\max }, \forall t \in \mathcal{T} \text{,}  \label{eq:moo-fomulate:e} \\
    &0 \leq y_{t}^{\mathrm{U}} \leq y_{\max }, \forall t \in \mathcal{T} \text{,}  \label{eq:moo-fomulate:f} \\
    &a_i(t) \cdot d_{i}^{h}(t) \leq d^{h}_{\text{max}}, \forall i \in \mathcal{N} \text{, } t \in \mathcal{T} \text{.} \label{eq:moo-fomulate:g}
\end{align}
\label{eq:moo-fomulate}
\end{subequations}
\noindent In~(\ref{eq:moo-fomulate:b}), the flight angle of the UAV at a given time slot is defined. Owing to speed limitations, the UAV cannot cover an infinite distance within a finite time slot, thereby restricting the flight distance in time slot $t$ as shown in~(\ref{eq:moo-fomulate:c}). Moreover, ~(\ref{eq:moo-fomulate:e}) and~(\ref{eq:moo-fomulate:f}) indicate that the UAV should be constrained within the designated target area. In addition,~(\ref{eq:moo-fomulate:d}) represents the discrete decision-making process when the UAV executes tasks. Additionally, to ensure communication quality,~(\ref{eq:moo-fomulate:g}) limits the maximum communication distance between the UAV and the GDs during each decision-making process.

\par The optimization problem involves decision variables $a_i(t)$ in~(\ref{eq:moo-fomulate:d}) and~(\ref{eq:moo-fomulate:g}) that are binary, which means that they can only take values of 0 or 1. As a result, the feasible set for this problem is discrete. In optimization theory, a problem is considered convex if the feasible set forms a convex region, where any convex combination of two feasible solutions also lies within the feasible set. However, due to the binary value of $a_i(t)$, the feasible set in our case does not satisfy this property. Specifically, the discrete and combinatorial nature of $a_i(t)$ introduces a non-convex feasible region, as the solutions cannot be interpolated between 0 and 1 in a continuous manner.

\par Furthermore, the objective functions $f_1$ and $f_2$ are dependent on these binary variables. Consequently, the overall objective Eq.~(\ref{eq:formu}) is non-convex because it is defined over a non-convex feasible set~\cite{lin2023pddqnlp}. This non-convexity implies that the optimization problem belongs to the class of mixed-integer nonlinear programming (MINLP), a known NP-hard problem. Therefore, the formulated problem is also an NP-hard problem. Moreover, in the considered scenario, various decisions rely on the instantaneous states of the constituent parts within the system, which makes it challenging to employ traditional optimization methods. 

\section{EMODRL-based Approach}
\label{sec:algorithm}

\par Based on the aforementioned analyses, the formulated CDECMOP is an NP-hard problem with an extremely complex solution space. For such multi-objective optimization problems (MOPs), the goal is to identify multiple non-dominated strategies. This approach ensures that the strategies remain effective even if the preferences change of the decision-maker in different usage scenarios. The main methods to achieve such strategies include multi-objective evolutionary algorithms and DRL methods that can accommodate multiple strategies. However, in the aforementioned scenario where the UAVs need to make frequent decisions and quick movements, multi-objective evolutionary algorithms typically suffer from issues such as premature convergence and local optima in dynamic environments when dealing with high-dimensional MOPs, leading to unacceptable non-dominated strategies~\cite{Xu2020}. Therefore, we propose a DRL algorithm based on an evolutionary framework to make decisions and control actions at each time slot within the system. However, as shown in Eq.~(\ref{eq:f1}), since the scheduling of tasks within the UAV involves complex long-term objectives that are independent of a single time slot, it is challenging to make decisions within each single time slot. Therefore, we first introduce a task scheduling strategy based on SA to simplify the actions, thereby transforming it into a more effective MOMDP.

\subsection{MOMDP Simplification and Formulation}

\par A multi-objective control problem can be formulated as a MOMDP, which is an extension of a standard Markov Decision Process (MDP) \cite{Xu2020}. The MOMDP is characterized by the tuple $\langle \mathcal{S}, \mathcal{A}, \mathcal{P}, \mathcal{R}, \bm{\gamma}, \mathcal{D}\rangle$. Specifically, $\mathcal{S}$ and $\mathcal{A}$ denote the state space and the action space, respectively, while $\mathcal{P}(s'|s,a)$ represents the state transition probability. Moreover, the vector of reward functions $\mathcal{R}$ is given by $[r_1,\dots,r_m]^{\top}$ with $r_i$ being a mapping from $\mathcal{S} \times \mathcal{A}$ to $\mathbb{R}$, $\bm{\gamma} = [\gamma_1, \dots,\gamma_m]^{\top}$ is the vector of discount factors with each $\gamma_i \in [0,1]$, and $\mathcal{D}$ denotes the initial state distribution, wherein $m$ indicates the number of objectives.

\par In a MOMDP, the expected cumulative reward of a policy $\pi: \mathcal{S} \rightarrow \mathcal{A}$ is defined as $\mathcal{J}^{\pi} = [\mathcal{J}_1^{\pi}, \ldots, \mathcal{J}_m^{\pi}]^{\top}$, where each $\mathcal{J}_i^\pi$ is defined in~\cite{Xu2020}. Furthermore, for a MOP with $M_p$ objectives, the objective function \( F_m \) is expressed as follows:  
\begin{equation}
    F_m(\pi) = \underset{\pi}{\text{Max}}(f_1^\pi, \ldots, f_m^\pi)\text{.}
\end{equation}
\noindent For this optimization problem, there exist two policies $\pi$ and $\pi'$ satisfying the following conditions: a) For any objective function $f_i$, satisfying $f_i^{\pi} \geq f_i^{\pi'} \text{,} 1 \leq i \leq M_p$. b) There exists at least one objective $f_j$ such that $f_j^{\pi} > f_j^{\pi'} \text{,} 1 \leq j \leq M_p$. Under these circumstances, the policy $\pi$ is said to Pareto dominate the policy $\pi'$, denoted as $F_m(\pi) > F_m(\pi')$. Thus, within the set of all feasible policies $\Pi$, there exists at least one policy $\pi_1 \in \Pi$ such that no other policy $\pi_2$ satisfies $F_m(\pi_2) > F_m(\pi_1)$. The policy $\pi_1$ is referred to as a non-dominated policy, characterized by not being dominated by any other policy. The set of all non-dominated policies constitutes the Pareto set. Visualizing the Pareto set in the objective space forms the Pareto front.  

\par As demonstrated in~\cite{Xu2020}, no single policy can simultaneously optimize every objective in a MOMDP. Instead, it is necessary to find a set of non-dominated policies, which corresponds to the aforementioned Pareto set. As such, once we obtain the Pareto set, we can identify the optimal policies within it according to any given preference criteria or conditions.
\subsubsection{Action Simplification}
\par As a control problem, CDECMOP typically treats its decision variables as potential actions within the MOMDP framework. Therefore, the action space of the MOMDP is expected to include scheduling decisions at each time slot. However, scheduling inherently represents a long-term problem that requires decomposition into individual time slots to enable decision-making. Moreover, scheduling does not occur in every time slot, necessitating the introduction of discrete decisions to determine whether scheduling is required at a given time slot.

\par Additionally, the number of tasks varies over time. Each time task quantities change, the model must be retrained before it can be applied. Furthermore, incorporating scheduling decisions into the MOMDP significantly increases the dimensionality of the discrete action space. When combined with actions for controlling flight and task offloading, this growth in action space hinders training efficiency and convergence performance.

\par To ensure both the generality and performance of the algorithm, we introduce a scheduling method to specifically handle the complex long-term scheduling problem, thereby simplifying the action space. There are many classical methods for task scheduling, such as shortest job first (SJF) and first-come-first-served (FCFS). However, these algorithms are usually not optimal and may be unstable for the considered problem~\cite{he2023}. Therefore, this paper proposes the use of the SA algorithm for task scheduling. The proposed scheduling method based on SA optimizes the long-term $D^s$ by improving decision-making for the task execution sequence. The specific implementation process of SA is detailed in Algorithm~\ref{alg4}. As can be seen, this method reduces one decision related to scheduling and the task index in the action space of the MOMDP, thereby effectively removing two types of actions from the action space. Since these reductions involve two discrete actions, where the first action has two decisions and the second corresponds to the task count in the current queue (approximately equal to the GDs count $N$), the overall dimensionality of the discrete action space is reduced by $2 \times N$. Furthermore, the potential action space includes both continuous and discrete actions, and the discrete actions can be approximated as continuous ones through this method. Consequently, this method significantly decreases errors introduced during the approximation process.

\begin{algorithm} 
    \caption{SA-based Task Scheduling Algorithm} 
    \label{alg4} 
    \begin{algorithmic}[1]
        \AlgorithmicInput{Task queue $Q$, maximum iterations $n_{\text{max}}$, cooling rate $\alpha$, temperature $T$;}
        \STATE Randomly initialize the current scheduling order $x$ as a permutation of $1,\dots,|Q|$;
        \WHILE{$T > T_{\text{min}}$ and $n_{\text{iter}} < n_{\text{max}}$}
            \FOR{$i = 1$ to $k$}
                \STATE Randomly swap elements in $x$ to generate $x'$;
                \STATE Calculate $\Delta C = \text{cost}(x') - \text{cost}(x)$;
                \IF{$\Delta C \geq 0$}
                    \STATE Calculate $P_c = e^{\left(-\frac{\Delta E}{T}\right)}$;
                    \IF{$P_c \geq \text{random}$}
                        \STATE Accept the new solution: $x \gets x'$;
                    \ENDIF
                \ELSE
                    \STATE Accept the new solution: $x \gets x'$;
                \ENDIF
            \ENDFOR
            \STATE Decrease temperature: $T \gets T \times \alpha$;
            \STATE Increment iteration: $n_{\text{iter}} \gets n_{\text{iter}} + 1$;
        \ENDWHILE
        \AlgorithmicOutput{Best schedule $x$.}
    \end{algorithmic} 
\end{algorithm}

\subsubsection{MOMDP Formulation}
\par The CDECMOP is a multi-objective control problem, thus it can be expressed as a MOMDP \cite{Xu2020}. To solve the formulated CDECMOP by using the EMODRL framework, we first transform the original optimization problem into a MOMDP. Specifically, the state space, action space, and reward functions are described as follows.
\begin{figure*}[!htb]
  \centering
  \includegraphics[width=1 \textwidth]{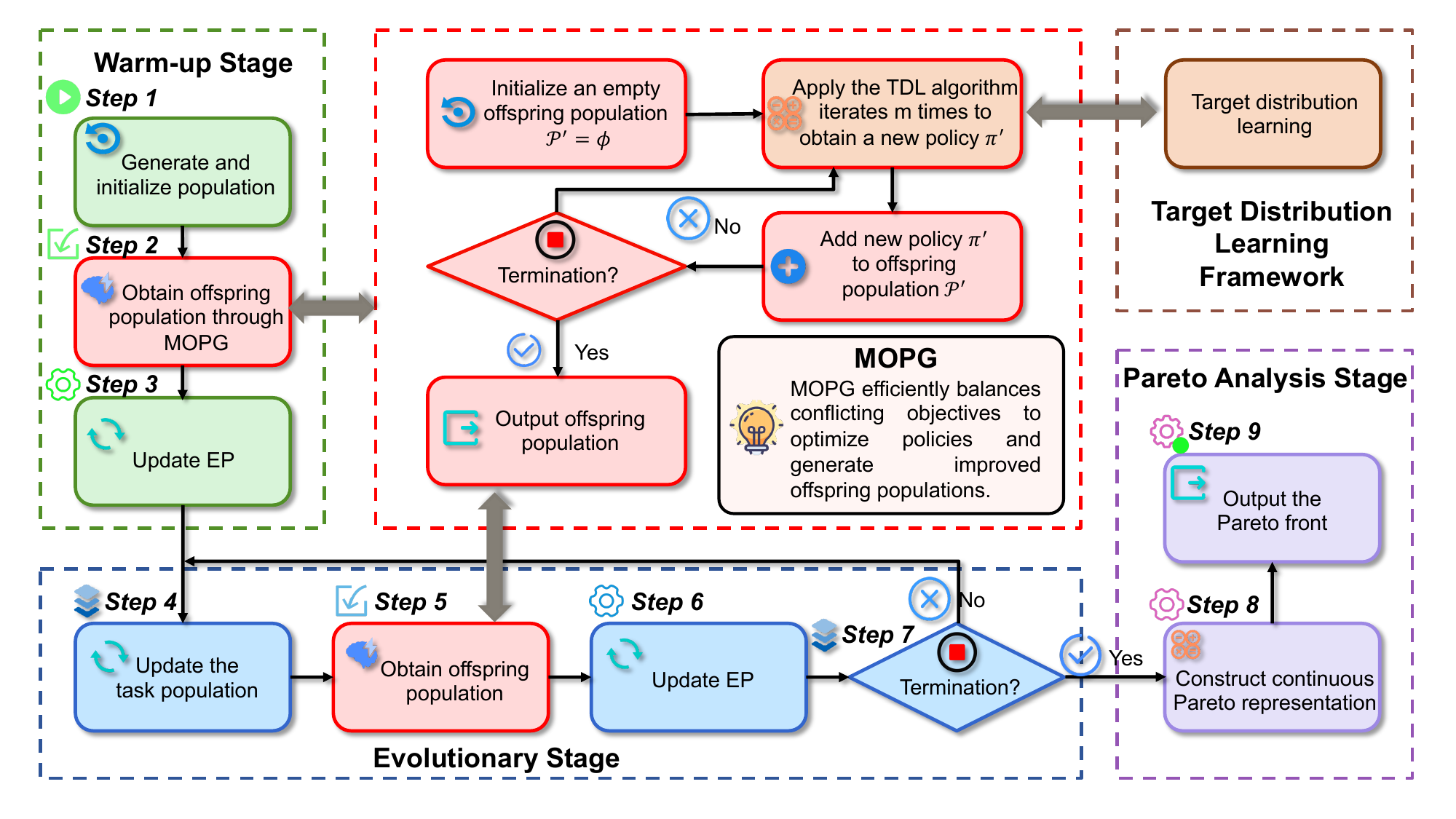}
  \caption{EMO-TDL-SA framework}
  \label{fig:Framework-MORL}
\end{figure*}

\begin{itemize}

\item \textit{State Space:} The design of the state space is crucial as it influences the convergence, rate of convergence, and overall performance of the DRL algorithm by governing the decision-making of the agent. In the considered scenario, the energy consumption and reward generation are jointly influenced by the interaction between the UAV and GDs, task transmission, and computation. Therefore, the information such as the timing and quantity of computational tasks generated by each GD, along with the real-time coordinate of the UAV, is critical for effective decision-making and performance optimization in the DRL algorithm. At time slot $t$, we define the state space as comprising the three main parts that are the current position of the UAV, the task queue within the UAV, and the status information of GDs. Accordingly, the state space is given as follows:
\begin{equation}
\displaystyle{S=\{s_{t}|s_{t}=(q_{t}^{\mathbf{U}}, N_{t}^{\mathbf{U}}, st_{t}^{\mathbf{GD}}),\forall t\in \mathcal{T}\}}
\text{,}
\end{equation}
\noindent where $q_{t}^{\mathbf{U}}=(x_{t}^{\mathbf{U}}, y_{t}^{\mathbf{U}}, H)$ represents the current coordinates of the UAV, $N_{t}^{\mathbf{U}}$ represents the size of the task queue in the UAV at time slot $t$, and $st_{t}^{\mathbf{GD}}$ is an $N$-dimensional vector, where the $i$-th element is denoted as $st_{t}^{GD_i} = (\lambda_{t}^{GD_i}, O_{t}^{GD_i})$. $\lambda_{t}^{GD_i}$ represents the start time of the task that exists at time slot $t$, and $O_{t}^{GD_i}$ denotes the size of the corresponding computational task.

\item \textit{Action Space:} The action space should offer possibilities to achieve the intended objectives and ensure proper accessibility to high-performance regions. Moreover, the action space should be as simple and efficient as possible to effectively reduce training difficulty and enhance algorithm performance. In the formulated MOMDP, the action space primarily consists of three parts, \textit{i.e.}, the direction of the movement of the UAV $\theta_t$, the distance of the movement of the UAV $d_t$, and the decision made by the UAV regarding tasks within the current time slot. As can be seen, the direction and distance of the movement of the UAV are continuous variables, while the decision is a discrete variable, making it an inapplicable setup for the MOMDP. To address this challenge, we transform the decision $\alpha_{t}$ of the UAV at each time slot into a continuous action space. Therefore, the designed action space is given as follows: 
\begin{equation}
    \displaystyle{A=\{a_{t}|a_{t}=(\theta_{t}, d_{t}, \alpha_{t}),\forall t\in \mathcal{T}\}} 
    \text{.}
\end{equation}

\item \textit{Reward Function:} The design of the reward function is an extremely important aspect in the application of DRL. By specifying and quantifying task objectives, rewards serve as a special language for effective communication between optimization goals and algorithms. The objective of autonomous UAV trajectory planning, GD task selection, and task scheduling is to minimize the task delay and the energy consumption of the UAV under given constraints. If any of the constraints specified in Eq.~(\ref{eq:moo-fomulate}) are not met, a penalty is imposed. As such, the reward function is designed as follows:
\begin{equation}
    r_{t}=
    \left\{\begin{array}
    {{l l}}{{-W,}}&{{\mathrm{if~}S_{t+1}=\Omega}}\\ 
    {(-D_{t}, -E_{t}),}&{{\mathrm{otherwise}}} \\
    \end{array}\right.
    \text{,}
\end{equation}
\noindent where $\Omega$ represents all the invalid states, and any state that fails to satisfy a specific constraint will be added to this set. Moreover, $W$ is a sufficiently large and reasonable positive number that ensures the satisfaction of these constraints during the learning process.
\end{itemize}

\subsection{The Proposed EMO-TDL-SA}
\begin{algorithm} [t]
    \caption{EMODRL Algorithm} 
    \label{alg1} 
    \begin{algorithmic}[1]
        \AlgorithmicInput{Number of population tasks $N_p$, total iterations in warm-up $M_w$, total generations $G$;}

        \textit{/* Starting The Warm-up Stage */}
        
        \STATE{Construct empty population $\mathcal{P} \gets \emptyset$, external pareto archive $\text{EP} \gets \emptyset$;}
        
        \STATE{Uniformly initialize a set of $N_p$ vectors $\mathcal{W} = \{w_1, \dots, w_{N_p}\}$;}

        \STATE{Randomly generate the initial policy set $\{\pi_1, \dots, \pi_{N_p}\}$;}
        \STATE{The initial task set is $\Gamma=\{\gamma^{\text{TS}}_1,\dots,\gamma^{\text{TS}}_{N_p}\}$,$\gamma^{\text{TS}}_i=\langle{w}_i,\pi_{i}\rangle$;}
        \STATE {$\mathcal{P^{\prime}} \gets \text{MOPG}(\Gamma, M_w)$;}
        \STATE {Add $\mathcal{P^{\prime}}$ to $\mathcal{P}$;}
        \STATE {Update EP by $\mathcal{P^{\prime}}$;}
        
        \textit{/* Starting The Evolutionary Stage */}
        \FOR{$i = 1,\dots,G$}
            \STATE {$\Gamma \gets \text{TaskUpdate($N_p$, }\mathcal{P}\text{, } w \text{);}$}
            \STATE {$\mathcal{P^{\prime}} \gets \text{MOPG}(\Gamma, M_w)$;}
            \STATE {Add $\mathcal{P^{\prime}}$ to $\mathcal{P}$;}
            \STATE {Update EP by $\mathcal{P^{\prime}}$;}
        \ENDFOR

        \textit{/* Starting The Pareto Analysis Stage */}
        \STATE {Perform k-means clustering on EP to obtain $\text{EP}^{\text{cluster}}$;}
        \STATE {Construct Pareto front through interpolation of $\text{EP}^{\text{cluster}}$;}

         \AlgorithmicOutput{Pareto front.}
    \end{algorithmic} 
\end{algorithm}

\subsubsection{Evolutionary Multi-objective Optimization Framework Overview}

\par The objective of the EMODRL algorithm is to find the Pareto policy set, and the proposed EMO-TDL-SA shares a similar framework with EMODRL. Furthermore, the framework of the EMO-TDL-SA is illustrated in Fig.~\ref{fig:Framework-MORL}, and the specific steps are outlined in Algorithm~\ref{alg1}. Specifically, the algorithm begins with the warm-up stage, wherein $N_p$ policies are randomly initialized, and each policy corresponds to a weight $w_i$ generated from a uniform distribution. Moreover, each pair of policy and weight $\langle \pi_i, w_i \rangle$ undergoes a fixed number of iterations of multi-objective policy gradient (MOPG) optimization, eventually forming the initial population.

\par The objective for a given policy $\pi$, weight vector $w$, and the number of objectives $m$ is to maximize the weighted reward $\mathcal{J}(\zeta,w)$ as follows:
\begin{equation}
    \mathcal{J}(\zeta,w)=\sum_{i=1}^{m}w_{i}J_{i}^{\pi} 
    \text{.}
\end{equation}

\par Therefore, the MOPG is employed to update the policy which is given as follows:
\begin{equation}
    \begin{aligned}
    \nabla_{\boldsymbol{\zeta}} \mathcal{J} (& \boldsymbol{\zeta}, \boldsymbol{\xi})=\sum_{i=1}^{m} \xi_{i} \nabla_{\boldsymbol{\zeta}} J_{i}(\boldsymbol{\zeta}) \\
    & =\mathbb{E}\left[\sum_{t=0}^{T} \boldsymbol{\xi}^{\top} \boldsymbol{A}^{\pi} \cfrac{\partial}{\partial \boldsymbol{\zeta}} \ln \pi_{\boldsymbol{\zeta}}\left(a_{t} \mid s_{t}\right)\right] \\
    & =\mathbb{E}\left[\sum_{t=0}^{T} A_{\boldsymbol{\xi}}^{\pi} \cfrac{\partial}{\partial \boldsymbol{\zeta}} \ln \pi_{\boldsymbol{\zeta}}\left(a_{t} \mid s_{t}\right)\right]\text{,}
    \end{aligned}
    \label{eq:MOPG}
\end{equation}
\noindent where $A_\xi^\pi = {\xi}A^\pi$ is the extended advantage function~\cite{Xu2020}. The MOPG mentioned can be implemented using various policy gradient algorithms in DRL, including PPO, soft actor-critic (SAC), \textit{etc}. This extension thereby allows these algorithms to be adapted as MOPG algorithms.

\par Subsequently, the algorithm enters the evolutionary stage. The policy population and the external Pareto set are optimized during this stage. For each weight vector, $N_p$ policies are selected, and each task is optimized using MOPG to generate the offspring population of the next generation. A maximum number of generations is set for the evolutionary stage, and when this limit is reached, the evolutionary stage terminates.

\par Finally, the algorithm proceeds to the Pareto analysis phase. With the external Pareto set obtained in the evolutionary stage, the original discrete Pareto policies are obtained. The Pareto analysis algorithm obtains a continuous Pareto representation through clustering and interpolation.

\subsubsection{Warm-up Stage}

\par In the warm-up stage, a set of $N_p$ random populations is initially generated, each consisting of a tuple of weight and policy. All policies within these populations share the same environment, including the state space, policy space, and reward function. However, due to different weight vectors, after undergoing one iteration of the policy gradient optimization designed for multiple objectives, a completely different initial population of $N_p$ policies is obtained. Once the initial population is secured, the external Pareto set is updated with this population.

\par Note that the warm-up stage allows the initial policies to move away from regions of low performance effectively.

\subsubsection{Evolutionary Stage}

\par In the evolutionary stage, the population is updated using the task selection algorithm, as depicted in Algorithm~\ref{alg3}.

\par For each weight vector $w_i$, the algorithm enumerates the policies within the current population and executes multiple rounds in the environment with $w_i$. Then, the task selection algorithm is employed to obtain the optimal policy $\pi'$ corresponding to $w_i$.

\par The evaluation of the Pareto metric includes indicators such as hypervolume and sparsity. The objective of the population update algorithm is to select the most pivotal populations to optimize the Pareto metric. In a practical implementation, a performance buffer is utilized. Within each performance buffer, all tasks are sorted based on their distance to the reference point $Z_{\text{ref}}$. Only the top $B_{\text{size}}$ tasks with the maximum distance are retained in each buffer. The optimal strategy is ultimately selected based on this process.

\par The extended Pareto set $EP$ retains the non-dominated policies generated across all evolutionary generations. Each newly generated population $P'$ in each generation updates $EP$. It iterates through all policies in $P'$ and incorporates any non-dominated policies into $EP$.

\par The evolutionary stage concludes upon reaching the predetermined number of evolutionary generations.

\begin{algorithm} [t]
    \caption{MOPG} 
    \label{alg2} 
    \begin{algorithmic}[1]
        \AlgorithmicInput{Task set $\Gamma$, total iterations $M$;}
        
        \STATE{Construct empty population $\mathcal{P^{\prime}} \gets \emptyset$;}
        
        \FOR{$\gamma^{\text{TS}}_i\langle{w}_i,\pi_{i}\rangle \in \Gamma$}
            \FOR{$\text{iter} = 1, \ldots, M$}
                \STATE{Train and update policy $\pi^{\prime}$ by policy gradient in Eq.~(\ref{eq:MOPG});}
            \ENDFOR
            \STATE{Add new policy $\pi'$ to offspring population $\mathcal{P}^{\prime}$;}
        \ENDFOR

    \AlgorithmicOutput{Offspring population $\mathcal{P^{\prime}}$.}
    \end{algorithmic} 
\end{algorithm}

\begin{algorithm} [t]
    \caption{TaskUpdate} 
    \label{alg3} 
    \begin{algorithmic}[1]
        \AlgorithmicInput{Number of tasks $N$, offspring population $\mathcal{P}$, weight vector $w$;}
        
        \STATE{Initialize empty task set ${\Gamma} \gets \emptyset$;}
        \FOR{$w_i \in w$}
            \STATE{Initialize candidate $pos \gets 1$;}
            \FOR{$j = 1, \dots, |\mathcal{P}|$}
                \STATE{Calculate the candidate value $\text{tar}_j = w_i \cdot F(\pi_j)$;}
                \IF{$\text{tar}_j$ outperforms  $w_i \cdot F(\pi_\text{pos}$)}
                    \STATE{Update $pos \gets j$;}
                \ENDIF
            \ENDFOR
            \STATE{Generate new task tuple $\gamma^{\text{TS}}_i = \langle w_i, \pi_{\text{pos}} \rangle$;}
            \STATE{Add task $\gamma^{\text{TS}}_i$ to $\Gamma$;}
        \ENDFOR

         \AlgorithmicOutput{Task set $\Gamma$.}
    \end{algorithmic} 
\end{algorithm}

\subsubsection{Pareto Analysis}

\par At the conclusion of the evolutionary stage, the $EP$ contains discrete Pareto-optimal policies. Through Pareto analysis, the Pareto front and the parameter structures of the policies on the Pareto front are determined.

\par To construct a smooth Pareto depiction, the reduced set of policies is grouped with the help of $k$-means clustering. Subsequently, linear interpolation is performed within each cluster to construct a continuous Pareto representation for the policies within the same cluster or series.
\subsubsection{TDL Framework}

\begin{figure*}[!htb]
  \centering
  \includegraphics[width=\textwidth]{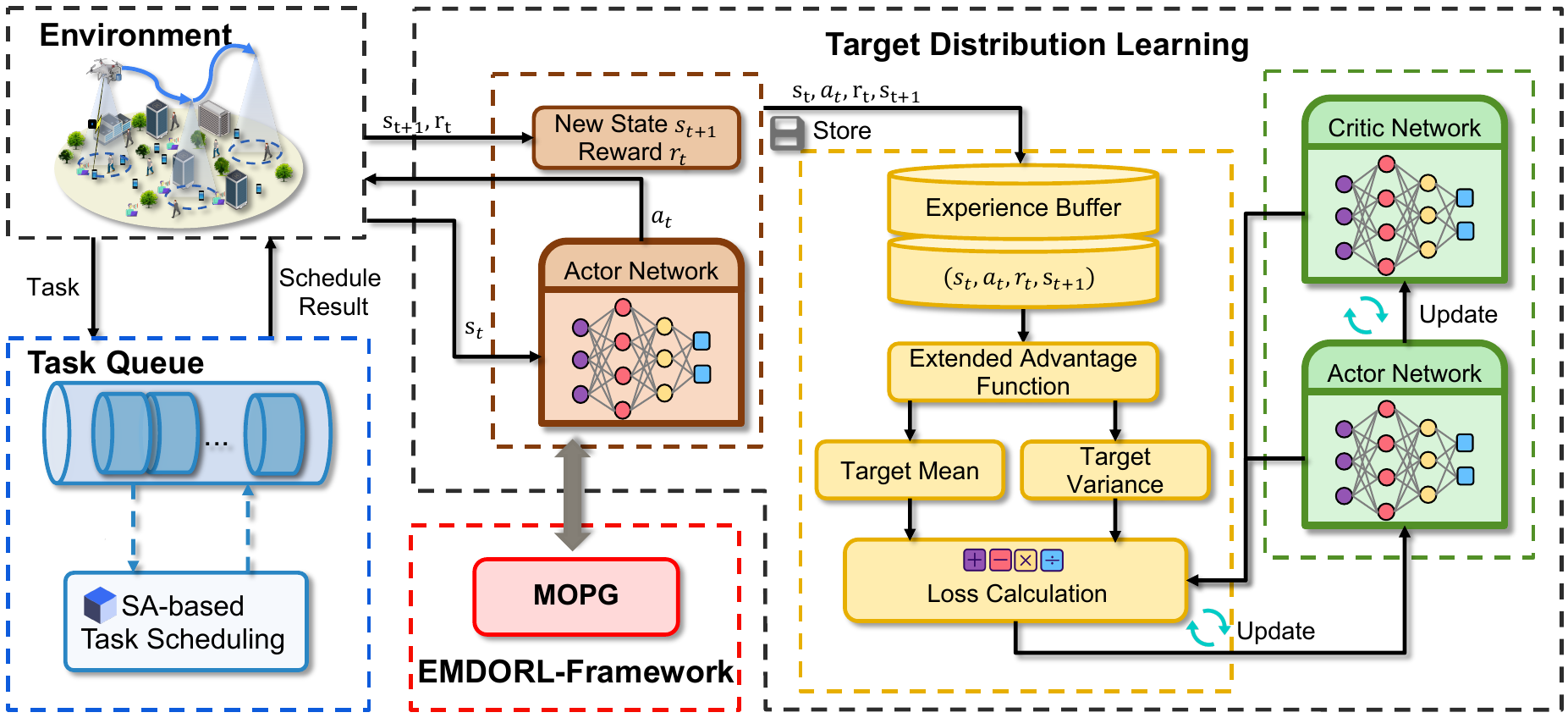}
  \caption{TDL Framework}
  \label{fig:Framework-TDL}
\end{figure*}

\par When using the MOPG training, instability may occur. The main reason is that when the variance of the neural network output distribution decreases, the policy gradient may become excessively large, thereby resulting in instability. To address this issue, we integrate the TDL algorithm~\cite{Zhang2020} into the EMODRL framework by extending it to a multi-objective version based on MOPG, as illustrated in Fig.~\ref{fig:Framework-TDL}. The detailed algorithm process is illustrated in Algorithm ~\ref{TDL}.

\par This method offers two main stability advantages. Firstly, it effectively limits the change of strategy in each iteration round, thereby making the training process more stable. Secondly, during each optimization step of one iteration, the neural network is optimized for a fixed goal, thereby making the optimization effect of this step less affected by the optimization parameters, which enhances the overall stability of the algorithm.

\par The TDL algorithm does not directly optimize the target $J(\zeta)$ but first introduces the parameters of the action distribution according to the advantage function, and trains the policy network. Simultaneously, TDL attempts to maximize a surrogate objective on state $s_t$, $L_{\{t,1\}}(\mu,\sigma)$ or with a probability to improve the old value function $L_{\{t,2\}}(\mu,\sigma)$, which can be given by
\begin{equation}
    L_{t, 1}(\mu, \sigma)=\mathbb{E}_{a \sim \mathcal{N}(\mu, \sigma)}\left[A^{\pi^{\text {old }}}_w\left(s_{t}, a\right)\right] 
    \text{,}
    \label{eq:Lt1}
\end{equation}
\begin{equation}
    L_{t, 2}(\mu, \sigma) = 
        \mathbb{E}_{a \sim \mathcal{N}(\mu, \sigma)} 
            \left[\mathbb{I}\left\{A_w^{\pi^{\mathrm{old}}}\left(s_{t}, a\right)>0\right\}
            \right]
    \text{,}
    \label{eq:Lt2} 
\end{equation}
\noindent while satisfying the constraint, which can be given by
\begin{equation}
    KL\left(\mathcal{N}\left(\cdot \mid \mu^{\text {old }}\left(s_{t}\right), \sigma^{\text{old }}\left(s_{t}\right)\right) \| \mathcal{N}(\cdot \mid \mu, \sigma)\right) \leq \delta
    \text{.}
    \label{eq:subject_to_L}
\end{equation}

\par To maximize Eq.~(\ref{eq:Lt1}) or Eq.~(\ref{eq:Lt2}) under the constraints in Eq.~(\ref{eq:subject_to_L}), following the extended advantage function described in the paper \cite{Zhang2020}, the variance and mean of the target distribution can be defined as follows:
\begin{equation}
    \begin{aligned}
        \hat{\sigma}_{t}^{2}=
        &\left(a_{t}-\mu^{\text{old }}\left(s_{t}\right)\right)^{2} \mathbb{I}\left\{\hat{A}^w_{t}>0\right\} \\
        &+\left(\sigma^{\text{old }}\left(s_{t}\right)\right)^{2} \mathbb{I}\left\{\hat{A}^w_{t} \leq 0\right\} \text{,}
    \end{aligned}
    \label{eq:target_sigma}
\end{equation}
\begin{equation}
    \hat{\mu}_{t}=\mu^{\text {old }}\left(s_{t}\right)+\operatorname{sign}\left(\hat{A}^w_{t}\right) \min \left(1, \frac{\sqrt{2 \alpha}}{\left\|y_{t}\right\|_{2}}\right) y_{t} \sigma^{\text {old }}\left(s_{t}\right) \text{,}
    \label{eq:target_mu}
\end{equation}
\noindent where $\operatorname{sign}(\cdot)$ is the sign function.

\par To prevent excessive variance updates that violate Eq.~(\ref{eq:subject_to_L}), the variance of the state can be updated as follows:
\begin{equation}
    \sigma_{\theta}\left(s_{t}\right)=\sigma^{1 /(\varphi+1)} \tilde{\sigma}_{\theta}\left(s_{t}\right)^{\varphi /(\varphi+1)}
    \text{,}
    \label{eq:update_sigma_theta}
\end{equation}
\noindent where $\varphi > 0$ is a hyperparameter, and $\sigma$ can be given by
\begin{equation}
    {\sigma}^{2}=\frac{1}{T} \sum_{t=1}^{T} \hat{\sigma}_{t}^{2}
    \text{.}
    \label{eq:update_sigma}
\end{equation}

\begin{algorithm} [t]
    \caption{Target Distribution Learning} 
    \label{TDL} 
    \begin{algorithmic}[1]
        \AlgorithmicInput{Task tuple $\gamma^{\text{TS}}\langle w,\pi\rangle$, number of timesteps $T$, minibatch size $M$, number of epochs $E$;}
        
        \STATE Interacting with the environment yields $T$ transitions ${(s_{t}, a_{t}, r_{t}, s_{t+1})}$;

        \STATE Calculate the extended advantage function;

        \STATE Calculate the target mean using Eq.~(\ref{eq:target_mu});
        \STATE Calculate the target variance using Eq.~(\ref{eq:target_sigma});
        
        \FOR{$i=1$ to $E \cdot T/M$}
            \STATE Sample $M$ transitions from the $T$ transitions;
            \STATE Update the policy network with the objective of minimizing the mean squared error between the target mean and variance;
            \STATE Update the critic network with the objective of minimizing the specified quantity;
        \ENDFOR
        \STATE Update $\sigma$ to $\hat{\sigma}$ according to the Eq.~(\ref{eq:update_sigma});
        \STATE Update $\sigma_{\theta}(\cdot)$ according to the Eq.~(\ref{eq:update_sigma_theta});
        
        \AlgorithmicOutput{The new policy network $\pi'$.}
    \end{algorithmic} 
\end{algorithm}
\subsection{Complexity Analysis}

\par The proposed EMO-TDL-SA comprises three main components, which are the warm-up stage, the evolutionary stage, and the Pareto analysis stage. We seek to determine the overall time complexity by analyzing each stage individually and then combining them.

\par \textbf{Warm-up Stage:} The warm-up stage involves training the model using Algorithm~\ref{alg2} for a single session. In this stage, each of the $N_p$ tasks is trained for $M_w$ iterations. Both the policy network and the critic network used in Algorithm~\ref{alg2} consist of $F$ fully connected layers, along with one input layer and one output layer. The number of neurons in the $l$-th layer of the neural network is defined as $n_l$. The time complexity for one forward and backward pass through the network is proportional to $\sum_{l=1}^{F+1} n_{l-1} \cdot n_l$~\cite{Chen2022a}. Considering $T$ time steps per iteration, the time complexity of the warm-up stage is $O( N_p \cdot M_w \cdot T \cdot \sum_{l=1}^{F+1} n_{l-1} \cdot n_l )$.

\par \textbf{Evolutionary Stage:} This stage runs over $G$ generations. In each generation, Algorithm~\ref{alg3} and Algorithm~\ref{alg2} are executed sequentially.

\par \textit{Algorithm~\ref{alg3}:} In Algorithm~\ref{alg3}, the objective is to find an optimal strategy for $N_p$ tasks, with each task having a population size of $|\mathcal{P}|$. The time complexity for this algorithm in one generation is $O( N_p \cdot |\mathcal{P}| )$.

\par \textit{Algorithm~\ref{alg2}:} For each task, Algorithm~\ref{alg2} trains the model over $M_w$ iterations. Similar to the warm-up stage, the time complexity for Algorithm~\ref{alg2} in one generation is $O( N_p \cdot M_w \cdot T \cdot \sum_{l=1}^{F+1} n_{l-1} \cdot n_l )$.

\par Considering that there are $G$ generations, the total time complexity for the evolutionary stage becomes $O( G \cdot [ N_p \cdot |\mathcal{P}| + N_p \cdot M_w \cdot T \cdot \sum_{l=1}^{F+1} n_{l-1} \cdot n_l ] )$. Typically, the term involving neural network training dominates, and thus we can approximate the time complexity of the evolutionary stage as $O( G \cdot N_p \cdot M_w \cdot T \cdot \sum_{l=1}^{F+1} n_{l-1} \cdot n_l )$.

\par \textbf{Pareto Analysis Stage:} The Pareto analysis stage involves processing the results obtained from the evolutionary stage to identify Pareto-optimal solutions. This typically involves sorting and comparing solutions, which have a time complexity significantly lower than that of the evolutionary stage and can be considered negligible~\cite{Xu2020}.

\par \textbf{Overall Complexity:} Combining the complexities of all stages, we find that the overall time complexity of Algorithm~\ref{alg1} is dominated by the evolutionary stage. Therefore, the total time complexity is $O( G \cdot N_p \cdot M_w \cdot T \cdot \sum_{l=1}^{F+1} n_{l-1} \cdot n_l )$.

\par This analysis indicates that the computational complexity is primarily determined by the number of generations $G$, the number of tasks $N_p$, the number of training iterations $M_w$, the time steps per iteration $T$, and the architecture of the neural networks involved (as characterized by $\sum_{l=1}^{F+1} n_{l-1} \cdot n_l$).

\section{Simulation Results}
\label{sec:simulation_results}

\par In this section, we conduct comprehensive simulations to evaluate the proposed EMO-TDL-SA. and show its convergence curves, Pareto front, and the obtained trajectory of UAV flight.

\begin{figure*}[!htb]
    \begin{minipage}[b]{0.48\textwidth}
        \centering
        \subfloat[]{
            \includegraphics[width=\textwidth]{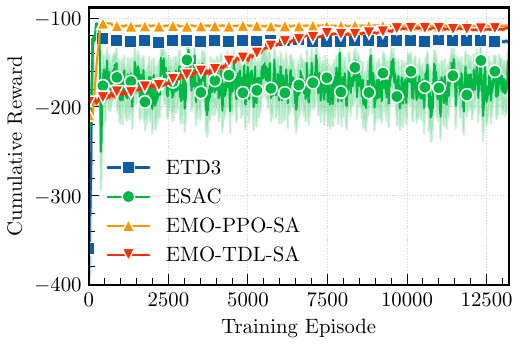}
            \label{fig:Result-Reward:CMP-1}
        }
    \end{minipage}%
    \hfill
    \begin{minipage}[b]{0.48\textwidth}
        \centering
        \subfloat[]{
            \includegraphics[width=\textwidth]{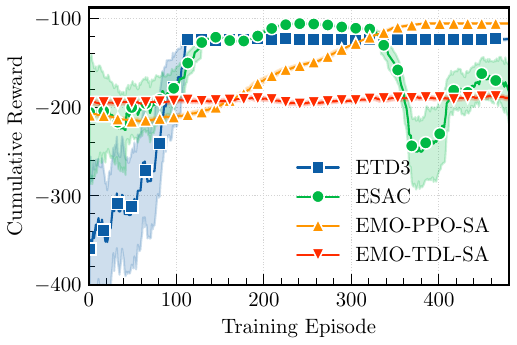}
            \label{fig:Result-Reward:CMP-2}
        }
    \end{minipage}   
    

    \caption{Convergence curves of various DRL algorithms. (a) Comparative convergence results of EMO-PPO-SA, EMO-TDL-SA, ETD3, and ESAC algorithms (highlighting the general convergence trends and performance differences). (b) Comparative convergence results of EMO-PPO-SA, EMO-TDL-SA, ETD3, and ESAC algorithms (highlighting the initial convergence in early training episodes). }
    \label{fig:Result-Reward}
\end{figure*}
\subsection{Simulation Settings}

\par The simulations are conducted using a Python simulator developed based on PyTorch 2.0. We consider the environment for the UAV-MEC system as follows.

\par The environment encompasses a region of $1000\text{ m} \times 1000 \text{ m}$. The flying altitude of the UAV is maintained at $H=100$ m, while the maximum flying speed is set at $30$ m/s. During each simulation, the horizontal position of the UAV is randomly determined. Additional simulation parameters are provided in Table~\ref{tab:simulation}.

\begin{table}[!t]
\caption{Simulation Settings \label{tab:simulation}}
\centering
\begin{tabular*}{\linewidth}{l l}
\hline
Simulation Parameters                                & Values                     \\ \hline
The rotor disc area $A$                              & 0.503 $m^2$~\cite{Wang2022a}   \\
The bandwidth of the UAV $B^{\text{UL}}$             & 10 $MHz$~\cite{Wang2022}            \\
The fuselage drag ratio $d_0$                        & 0.6~\cite{Wang2022}            \\
The blade profile power $P_1$                        & 79.8563 $W$~\cite{Wang2022a}   \\
The induced power $P_2$                              & 88.6279 $W$~\cite{Wang2022a}   \\
The rotor solidity $s$                               & 0.05~\cite{Wang2022}           \\
The velocity at the tip of the rotor blade $v_{\text{tip}}$    & 120 $m/s$~\cite{Wang2022a}     \\
The average velocity induced by the rotor $v_{0}$              & 4.03 $m/s$~\cite{Wang2022a}    \\
The maximal azimuth angle $\theta^{\text{max}}$      & $\pi/4$~\cite{Wang2022}        \\
The losses for LoS $\eta_{\text{LoS}}$               & 0.1 $dB$~\cite{AlHourani2014}  \\
The losses for non-LoS $\eta_{\text{NLoS}}$          & 21 $dB$~\cite{AlHourani2014}   \\ \hline
The total generations $G$                            & 500                        \\
The learning rate $l_r$                                  & $10^{-4}$                  \\ 
The number of warm-up iterations $M_w$               & 60                         \\
The number of tasks $N_p$                            & 10                         \\
The replay buffer size                               & $10^5$                     \\ \hline
\end{tabular*}
\end{table}

\par To compare, this work introduces several types of algorithms and methods for evaluation as follows:

\begin{itemize}

\item \textit{Naive Greedy Method}: The UAV trajectory design uses simple paths, such as random walk, circular, and spiral trajectories, to ensure coverage of all GDs. For task processing, a greedy strategy is employed to prioritize the tasks from the nearest GDs for execution.

\item \textit{ESAC}: ESAC is an evolutionary multi-objective SAC algorithm, an extension of the single-objective SAC algorithm. ESAC implements the standard EMORL as described in~\cite{Xu2020}. The algorithm uses the SAC algorithm to enable MOPG and produces Pareto policies. The implementation of ESAC aims to validate the performance of our proposed algorithm.

\item \textit{ETD3}: ETD3 is another implementation of EMORL, which extends the twin-delayed deep deterministic policy gradient (TD3) algorithm to handle multiple objectives. The algorithm is the same as the ETD3 used in \cite{Song2022}. Similarly, we implement ETD3 for performance comparison.

\item \textit{EMO-PPO-SA}: This is the basic algorithm based on our proposed framework, with MOPG implemented using PPO. The algorithm also employed SA to simplify the action space, aiming to validate the performance of our TDL algorithm, which is improved based on PPO.

\item \textit{EMO-TDL-SA}: This is the optimized algorithm derived by enhancing PPO through the introduction of Algorithm~\ref{TDL}, which is the main algorithm proposed in this paper. To verify the effectiveness of our proposed scheduling strategies, we conduct simulations using different scheduling approaches, which are FCFS, SJF, and Priority Scheduling (PS). The corresponding algorithms are named EMO-TDL-FCFS, EMO-TDL-SJF, and EMO-TDL-PS, respectively.

\end{itemize}


\subsection{Convergence Analysis}

\begin{figure}[htp]
    \centering
    \includegraphics[scale=1]{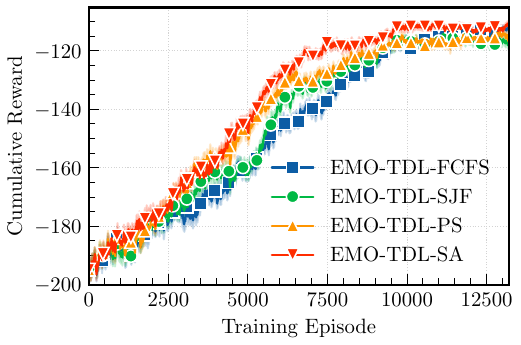}
    \caption{Convergence trend for EMO-TDL with FCFS, SJF, SA, and PS scheduling strategies.}
    
    \label{fig:TDL-Result-Reward}
\end{figure}
\par In general, DRL-based models must converge during training before they can be deployed and put into use. Even in dynamically changing real-world environments, the model typically requires retraining and fine-tuning before deployment. Fig.~\ref{fig:Result-Reward}(a) compares the convergence curves of four DRL-related algorithms, \textit{i.e.,} ESAC, ETD3, EMO-TDL-SA, and EMO-PPO-SA. Apart from ESAC, all eventually converge, though ETD3 converges to a slightly lower value than EMO-PPO-SA and EMO-TDL-SA. For more details, Fig.~\ref{fig:Result-Reward}(b) shows the convergence results in the early convergence phase. As can be seen, ESAC initially converges but eventually falls into a suboptimal outcome with oscillations, failing to achieve stable convergence. Additionally, while EMO-TDL-SA takes the longest time to converge among the four algorithms, it ultimately stabilizes at the best result. 

\par Moreover, we compare the convergence of EMO-TDL-SA with that of various scheduling strategies. Fig.~\ref{fig:TDL-Result-Reward} shows the convergence of the EMO-TDL with different scheduling strategies, respectively. As can be seen, each strategy eventually converges. However, EMO-TDL-SA converges slightly faster and achieves a better final value compared to the other three strategies. This indicates that different scheduling methods affect the final convergence outcome, and SA-based scheduling helps EMO-TDL perform the best. 

\begin{figure}[htp]
    \centering
    \includegraphics[scale=1]{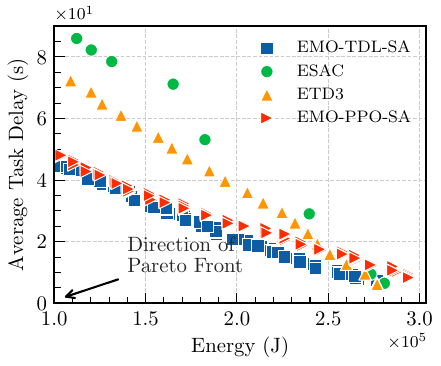}
    \caption{Solution distributions obtained by different algorithms.}
    
    \label{fig:pareto}
\end{figure}

\subsection{Performance Comparison}

\par As shown in Fig.~\ref{fig:pareto}, we present the trade-offs between the two considered objectives obtained by multiple EMODRL algorithms. Each of these algorithms achieves a set of Pareto-optimal strategies, demonstrating the effectiveness of our EMODRL framework in yielding various strategies that balance the objectives. Furthermore, from the direction of the Pareto front shown in the figure, it can be observed that the strategies obtained by EMO-TDL-SA consistently dominate almost all other strategies, thereby achieving the best Pareto policies.

\par
\begin{figure}[htp]
    \centering
    \includegraphics[scale=0.8]{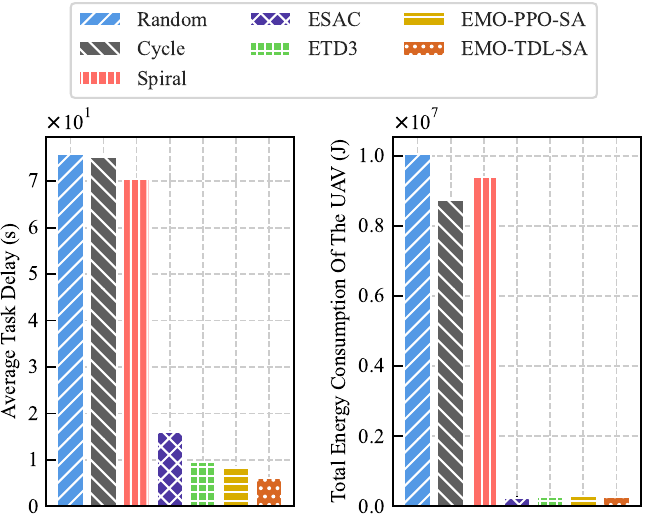}
    \caption{The optimization objective values of different approaches.}
    
    \label{fig:result_bar}
\end{figure}

 Moreover, to intuitively compare the effectiveness of the EMODRL methods with other naive methods, we select a balanced strategy from the Pareto-optimal strategies of each EMODRL method. We then obtain the numerical results and compare them with three different flight trajectories where nearby nodes are selected greedily. Additionally, the numerical results of all algorithms are presented in Fig.~\ref{fig:result_bar}. As can be seen, the numerical results of the four EMODRL methods yield lower energy consumption and task delay than those of the three specific trajectories combined with a greedy strategy. This is because the three algorithms are EMODRL methods, which can better balance each optimization objective and find an optimal action at each step. Among them, ESAC achieved the smallest total energy consumption of the UAV, and EMO-TDL-SA achieved the smallest average task delay. However, the ESAC algorithm obtained the largest average task delay among the four EMODRL methods, several times greater than the average task delay of the other three methods, indicating excessive task delay for task completion. Additionally, the differences in energy consumption of the UAV among ESAC, ETD3, EMO-PPO-SA, and EMO-TDL-SA are negligible relative to their total energy consumption of the UAV. Consequently, all four methods achieve nearly the same energy consumption. However, EMO-TDL-SA achieves significantly lower task delay with comparatively low energy consumption of the UAV. Therefore, we can conclude that EMO-TDL-SA is the most suitable approach for the considered scenario. 

\par
\begin{figure}[htp]
    \includegraphics[width=3.2 in]{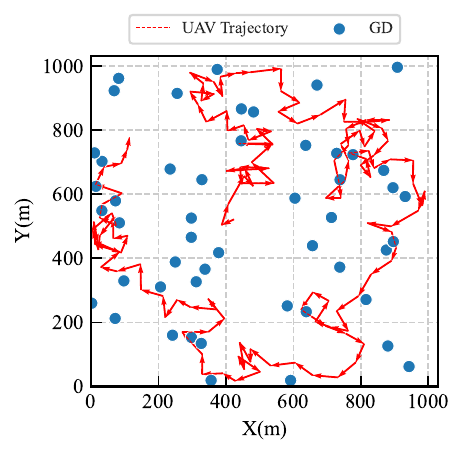}
    \caption{The UAV trajectory under the balanced strategy of EMO-TDL-SA.}
    
    \label{fig:trace}
\end{figure}

 Additionally, we provide the UAV trajectory for EMO-TDL-SA under the balanced strategy in Fig.~\ref{fig:trace}. As can be seen, the trajectory demonstrates excellent coverage of the target area, thus ensuring that all regions of interest are effectively monitored. Meanwhile, the method achieves high energy efficiency, as the trajectory is optimized to minimize unnecessary movements and energy consumption of the UAV. This balance between coverage and energy utilization highlights the effectiveness of the proposed approach in addressing the dual objectives of comprehensive monitoring and resource conservation.


\section{Conclusion}
\label{sec:conclusion}
\par This paper has investigated a UAV-assisted MEC system in LAE. Specifically, a UAV-carried edge server has been used to facilitate task offloading for GDs. In this system, we have formulated the CDECMOP to minimize the energy consumption of the UAV and the total delay of tasks for the GDs. However, due to the complex long-term scheduling problem, it is difficult to directly control the system within each time slot. Therefore, we have proposed the SA-based scheduling method to reduce the action space and address this issue. In this way, we have formulated the entire problem as a MOMDP. Then, we have developed the EMO-TDL-SA for a more stable acquisition of different strategies, which further adapts to various scenarios. Simulation analyses have revealed that EMO-TDL-SA demonstrates superior convergence stability and performance metrics compared to existing DRL algorithms. Through extensive evaluation of final strategies, EMO-TDL-SA not only surpassed various baseline algorithms but also successfully generated superior Pareto-optimal policies, thus highlighting its robust optimization capabilities. 

\bibliographystyle{IEEEtran}

\vspace{11pt}
\vfill
\end{document}